\documentclass[runningheads]{llncs}

\usepackage{eccv}

\usepackage{eccvabbrv}

\usepackage{graphicx}
\usepackage{booktabs}
\usepackage{multicol}
\usepackage{multirow}
\usepackage{import}

\usepackage{amsmath,amsfonts,bm}

\def\eqref#1{equation~\ref{#1}}

\def\1{\bm{1}}

\def\vepsilon{{\bm{\epsilon}}}

\def\vc{{\bm{c}}}

\def\vp{{\bm{p}}}

\def\vx{{\bm{x}}}
\def\vy{{\bm{y}}}

\DeclareMathAlphabet{\mathsfit}{\encodingdefault}{\sfdefault}{m}{sl}
\SetMathAlphabet{\mathsfit}{bold}{\encodingdefault}{\sfdefault}{bx}{n}

\def\gL{{\mathcal{L}}}

\def\gN{{\mathcal{N}}}

\def\sE{{\mathbb{E}}}

\newcommand{\KL}{\mathbb{KL}}

\usepackage[accsupp]{axessibility}  

\newcommand{\ASP}{\texttt{ASP}}
\newcommand{\TIP}{\texttt{TIP}}
\newcommand{\TSP}{\texttt{TSP}}

\usepackage{hyperref}

\usepackage{orcidlink}

\begin{document}

\title{Few-Shot Class Incremental Learning with Attention-Aware Self-Adaptive Prompt} 

\titlerunning{FSCIL with Attention-aware Self-adaptive Prompt}

\author{Chenxi Liu  \and
Zhenyi Wang  \and
Tianyi Xiong  \and 
Ruibo Chen  \and 
Yihan Wu  \and 
Junfeng Guo  \and 
Heng Huang }

\authorrunning{C.~Liu et al.}

\institute{University of Maryland, College Park, MD 20742, USA \\
\email{\{cxliu539, heng\}@umd.edu}}

\maketitle

\begin{abstract}
  Few-Shot Class-Incremental Learning (FSCIL) models aim to incrementally learn new classes with scarce samples while preserving knowledge of old ones. Existing FSCIL methods usually fine-tune the entire backbone, leading to overfitting and hindering the potential to learn new classes. On the other hand, recent prompt-based CIL approaches alleviate forgetting by training prompts with sufficient data in each task. In this work, we propose a novel framework named \textbf{A}ttention-aware \textbf{S}elf-adaptive \textbf{P}rompt (\ASP{}). \ASP{} encourages task-invariant prompts to capture shared knowledge by reducing specific information from the attention aspect. Additionally, self-adaptive task-specific prompts in \ASP{} provide specific information and transfer knowledge from old classes to new classes with an \textit{Information Bottleneck} learning objective. In summary, \ASP{} prevents overfitting on base task and does not require enormous data in few-shot incremental tasks. Extensive experiments on three benchmark datasets validate that \ASP{} consistently outperforms state-of-the-art FSCIL and prompt-based CIL methods in terms of both learning new classes and mitigating forgetting. Source code is available at \url{https://github.com/DawnLIU35/FSCIL-ASP}.

\end{abstract}

\section{Introduction}
\label{introduction}

As the world keeps changing over time, the data in the world also continually changes. Thus, there is a need for machine learning models to follow the data change and continually learn new classes while preserving the knowledge learned from previous data, which is called as Class-Incremental Learning (CIL)~\cite{van2019three, Wu_2019_CVPR, masana2022class, wang2023metamix}. The main challenge of CIL is the catastrophic forgetting problem~\cite{french1999catastrophic, Forgetting_Survey_2023}: a model forgets the previous knowledge after training on new tasks, as the data from old tasks are not fully available due to reasons like limited storage space or privacy issues~\cite{de2021continual,wang2023federated}. While many CIL methods assume a model can continually train on new class data with sufficient samples~\cite{li2017learning, goswami2023fecam,liudeja}, this assumption does not hold in many real-world applications. For instance, in the scenario of an intelligent medical decision system tracking physiological signals, new patients with limited data must be learned without discarding knowledge from existing patient data~\cite{sun2023few}. This task of continually learning new classes with limited data is called Few-Shot Class-Incremental Learning (FSCIL)~\cite{tian2024survey, Tao_2020_CVPR}. FSCIL typically involves training a base model using a set of base classes with sufficient samples, subsequently utilizing the knowledge learned from base classes to facilitate the incremental learning of new classes with limited samples. Beyond the challenge of catastrophic forgetting, FSCIL also triggers overfitting on limited training samples, making it harder for a machine model to learn new classes.

Various works~\cite{tian2024survey} have been proposed to address the FSCIL scenario. Some of them focus on enhancing base models' ability to generalize across newly encountered few-shot classes~\cite{shi2021overcoming, Zhu_2021_CVPR, chi2022metafscil}, while others aim to find a better strategy to incrementally train on new tasks with limited data~\cite{Kukleva_2021_ICCV, cheraghian2021semantic, dong2021few, wang2022meta}. However, most existing works fine-tune all the parameters in the base model, which leads to overfitting on base classes and hinders the transferability to new classes. On the other hand, recent prompt-based CIL methods~\cite{wang2022learning, wang2022dualprompt, smith2023coda} leverage the inherent generalization ability of pre-trained Vision Transformer (ViT)~\cite{dosovitskiy2020image} by fixing the backbone parameters and only training a few new parameters called prompts~\cite{lester-etal-2021-power, wei2022chain}. They usually learn task-specific prompts via a key-query mechanism and store the knowledge of seen tasks in a specialized prompt pool. In this way, they preserve the knowledge of old tasks without necessitating a rehearsal buffer to store old data samples. Nonetheless, to train the task-specific prompts, prompt-based approaches require sufficient data samples from new tasks, which are not available in few-shot incremental tasks.

In this paper, we propose a novel Attention-Aware Self-Adaptive Prompt (\ASP{}) framework to overcome the shortcomings of existing FSCIL and prompt-based CIL methods under the FSCIL setting. Aiming to facilitate the continual learning of new classes with limited data, \ASP{} leverages the inherent generalization capability of the pre-trained ViT and the knowledge learned from sufficient base classes. Specifically, \ASP{} fixes the ViT backbone and introduces prompts between attention blocks for adapting to FSCIL tasks, where prompts are decomposed into attention-aware task-invariant prompts (\TIP{}) and self-adaptive task-specific prompts (\TSP{}). The attention blocks pay the same attention to each \TIP{} regardless of the task, encouraging \TIP{} to only contain task-invariant information that can be universally used for both base classes and new classes. Unlike the previous key-query mechanics, \ASP{} uses a prompt encoder to convert input images to prompt features. Inspired by the \textit{Information Bottleneck} (IB) theory~\cite{alemi2017deep}, \ASP{} guides the prompt encoder to generate prompt features that have a strong correlation with semantic information and a weak correlation with extraneous information contained in images. To further improve the generalization ability, \ASP{} aggregates prompt features over the training set to prevent overfitting on a single input image. For a specific input image, the corresponding \TSP{} is composed of the average prompt features and its own prompt features. Consequently, \ASP{} avoids fine-tuning the entire backbone to alleviate overfitting and circumvents the need of sufficient data to train new \TSP{} for new classes. Lastly, to further enhance model discrimination, a similarity-based loss is used to cluster feature vectors to their class centers, where class centers are estimated by the anchor samples during training.


Overall, \textbf{our contributions} can be summarized as follows:
\begin{itemize}
    \item[$\bullet$] We propose \ASP{}, an innovative prompt-based methodology to address both the overfitting challenge in existing FSCIL methods and the data-hungry drawback in existing prompt-based methods under the FSCIL scenario.
    
    \item[$\bullet$] We design attention-aware \TIP{} and self-adaptive \TSP{} to transfer knowledge from base classes to new classes and alleviate forgetting on old classes.

    \item[$\bullet$] Extensive experiments on three benchmark datasets demonstrate that \ASP{} substantially outperforms SOTA FSCIL and prompt-based CIL methods in both learning new classes and preserving performance on old classes.
\end{itemize}
\section{Related Work}
\label{related_work}

\subsection{Class-Incremental Learning}

\textbf{Non-prompt-based approaches:} In general, there are three different settings of incremental learning: task-, domain, and class-incremental learning(TIL, DIL, and CIL)~\cite{van2019three}. Among all, CIL is considered to be the most challenging scenario~\cite{tian2024survey}, which is required to learn new classes without forgetting old ones. In current CIL works, there are three main directions. The most effective one is rehearsal methods~\cite{rebuffi2017icarl, aljundi2019gradient, rolnick2019experience, wang2023distributionally, wang2024a}, which build a rehearsal buffer to store samples from previous tasks. The second direction is to find out important parameters for the current task and prevent them from changing over incremental tasks~\cite{aljundi2018memory, kirkpatrick2017overcoming, zenke2017continual}. In addition, a lot of works utilize knowledge distillation to preserve the knowledge of previous tasks to overcome forgetting~\cite{hinton2015distilling, li2017learning, rebuffi2017icarl}. Recently, some rehearsal-free methods~\cite{lomonaco2020rehearsal, wang2022dualprompt} have caught people's attention, as rehearsal samples are not always allowed to be stored in real-world scenarios~\cite{wang2022learning}. Notably, \ASP{} also doesn't need a rehearsal buffer to store any data samples. Typically, CIL methods require enough training data in each incremental task to learn new classes, which is not available under the FSCIL scenario.



\noindent\textbf{Prompt-based approaches:} Prompt-based methods~\cite{lester-etal-2021-power, wei2022chain} are first proposed for Natural Language Processing tasks, which can better utilize the pre-trained knowledge for downstream tasks. The basic idea is fixing the backbone parameters and only fine-tuning a few new parameters (prompts) prepend to input text or images~\cite{jia2022visual}. Recently, prompt-based CIL approaches using ViT backbones achieve significant performance in both learning new classes and preventing catastrophic forgetting. L2P~\cite{wang2022learning} first proposes to use a key-query mechanism to select task-specific prompts from a prompt pool. DualP~\cite{wang2022dualprompt} adds task-invariant prompts to capture the shared information among tasks. But their task-invariant prompts still provide much task-specific information and their task-specific prompts require sufficient training data in incremental tasks. This would result in significant overfitting to new few-shot incremental tasks because of the limited availability of new task data, ultimately causing a decline in performance for FSCIL. Subsequently, CodaP~\cite{smith2023coda} proposes to train the prompt pool and selection mechanism in an end-to-end manner. The most recent HideP~\cite{wang2023hierarchical} decomposes CIL into hierarchical components and optimizes them respectively. However, all existing prompt-based CIL methods are not suitable under the FSCIL scenario, as they all need sufficient data samples in incremental tasks to capture task-specific knowledge and store them in prompts. Reversely, \ASP{} does not need to train new task-specific prompts for new tasks and thus works well in few-shot incremental tasks.

\subsection{Few-Shot Class-Incremental Learning}

Few-Shot Class-Incremental Learning (FSCIL) is even more challenging than normal CIL, as it needs to incrementally learn new classes with limited labeled data~\cite{tian2024survey}. TOPIC~\cite{tao2020few} first proposed the FSCIL task, which is required to first obtain a base model trained on sufficient base classes, then incrementally learn some new tasks with limited new class data. Current FSCIL methods can be roughly divided into two groups. The first group aims to utilize base classes to train a generalized backbone that can be transferred to few-shot incremental tasks~\cite{shi2021overcoming, Zhu_2021_CVPR, chi2022metafscil}. The trained backbone is usually kept frozen during the following few-shot incremental tasks~\cite{song2023learning, Zhang_2021_CVPR, peng2022few}. The second group focuses on the strategy of incrementally learning few-shot new classes without overfitting~\cite{Kukleva_2021_ICCV, cheraghian2021semantic, dong2021few}. These FSCIL methods usually fine-tune all parameters in the backbone when training on base classes, which leads to overfitting on base classes and hinders the transferability to new classes. In contrast, \ASP{} fixes the pre-trained backbone and stores task-invariant knowledge in prompts to overcome this issue.
On the other hand, some recent studies~\cite{d2023multimodal,yoon2023image,huang2024learning} utilize OpenAI CLIP~\cite{radford2021learning} to handle multi-modal inputs, requiring alignment of features between image and text. Conversely, \ASP{} employs a ViT backbone with single-modal inputs, simplifying the training process and enhancing the model’s focus on visual features alone.

\section{Preliminaries}
\label{preliminaries}


\textbf{Few-shot Class-incremental Learning} aims to learn a sequence of tasks $t$ using their respective data $\mathcal{D}_0, ..., \mathcal{D}_T$. When learning on task $t$, data from previous tasks $0, ...,t-1$ are totally unavailable or only partially available, and the model is required to perform well on all seen tasks $0, ...,t$. The training data in task $t$ is represented as $\mathcal{D}_{t} = \{(\vx_{t,i}, y_{t,i})\}_{i=1}^{N_t}$, where $ N_t = |\mathcal{D}_{t}|$ denotes the size of $\mathcal{D}_t$, $\vx_{t, i} \in \mathcal{\bm{X}}_t$  and $y_{t, i} \in \mathcal{Y}_t$ represent the sample and label, respectively.  The training label spaces between different tasks are disjoint, i.e., for any task $t,t' \in [0,T]$ and $t \neq t'$, $\mathcal{Y}_t \cap \mathcal{Y}_{t'} = \varnothing$. The first task has sufficient training data $\mathcal{D}_{0}$ and is called the base task, while the following incremental tasks can be denoted as $N$-way $K$-shot classification tasks, i.e., $N$ classes for each task and $K$ samples for each class. A FSCIL model can be decoupled into a backbone $f_\theta$ parameterized by $\theta$, and a linear classifier $h_\psi$ parameterized by $\psi$. For an input test data $\vx$ drawn from all seen tasks, the model tries to predict $y = h_\psi(f_\theta(\vx))$ which matches the class label.



\noindent\textbf{Prompt-based Approaches} for vision tasks typically use a pre-trained vision transformer (ViT)~\cite{dosovitskiy2020image} as the backbone $f_\theta$, and the parameter $\theta$ are typically frozen during training to maintain the generalization ability obtained from the pre-training.  A ViT model contains multiple multi-head self-attention (MSA) layers, and we denote the input of the $l^{th}$ MSA layer as $\boldsymbol{h}^l \in \mathbb{R}^{L_{\boldsymbol{h}} \times D}$, then the output of this layer is given as:
\begin{small}
\begin{equation}
\begin{split}
    {\rm{MSA}}(\boldsymbol{h}^l) = {\rm{Concat}} (h_1^l, ..., h_m^l) W^O, \;  \text{where}\; h_i^l = {\rm{Attention}} (\boldsymbol{h}^l W^{Q}_{i}, \boldsymbol{h}^l W^{K}_{i}, \boldsymbol{h}^l W^{V}_{i})\\
\end{split}
\end{equation}
\begin{equation}
\begin{split}
    {\rm{Attention}} (Q, K, V) = {\rm{softmax}} (\frac{Q K^T}{\sqrt{d_k}}) V 
\end{split}
\end{equation}
\end{small}

\noindent 
where $W^O, W^{Q}_{i}, W^{K}_{i}, W^{V}_{i}$ are projection matrices, $m$ is the number of attention heads and $d_k$ is a scaling factor. Among prompt-based approaches, Prompt Tuning (ProT)~\cite{lester-etal-2021-power, jia2022visual} is one of the most commonly used techniques which introduce a few trainable parameters $\boldsymbol{p}^l \in \mathbb{R}^{L_{\boldsymbol{p}} \times D}$ as prompts for the $l^{th}$ layer, and these prompts are prepend to $\boldsymbol{h}^l$:
\begin{small}
\begin{equation}
\begin{split}
    f_{{\rm{ProT}}} (\boldsymbol{p}^l, \boldsymbol{h}^l) = {\rm{MSA}}([\boldsymbol{p}^l;\boldsymbol{h}^l])
\end{split}
\end{equation}
\end{small}

\noindent
where $[\cdot ~; \cdot]$ denotes the concatenation operation along the dimension of sequence length. Before the first layer of ViT, an input image is first split as a few patches and transformed into a sequence-like representation $\vx^e \in \mathbb{R}^{L_{\vx} \times D}$. For image classification tasks~\cite{jia2022visual}, a class token $\boldsymbol{cls} \in \mathbb{R}^{1 \times D}$ is prepend and the visual prompts are prepend to form the input of ViT blocks:
\begin{small}
\begin{equation}
\begin{split}
    {\vx^p} = [\boldsymbol{cls} ; \boldsymbol{p}^0 ; \boldsymbol{\vx^e}]\\
\end{split}
\end{equation}
\end{small}



\noindent\textbf{Prototypical Network}~\cite{snell2017prototypical} is a widely used approach for few-shot learning problems. It calculates the mean features $c_k$ of a class $k$ and uses it as the class prototype, i.e., ${\vc_k} = \frac{1}{N_k} \sum_{y_i=k} f( \vx_i)$; 
where $N_k$ is the number of samples in class $k$, and the output feature of the $\boldsymbol{cls}$ token is used as the image embedding. For a classification task with $K$ classes, $W=[\vc_0, \vc_1,...,\vc_K]$ is used as the linear classifier, and an input sample is classified via the softmax probability with class prototypes: $P(y=k|\vx) \propto \vc_k^T f_{\theta,\boldsymbol{p}}(\vx)$. Following works~\cite{wang2023few, zhou2022forward}, new class prototypes are append to $W$ to perform classification over all seen classes.

\section{Methodology}
\label{methodology}

\begin{figure*}[t]
    \centering
    \includegraphics[width=1\textwidth]{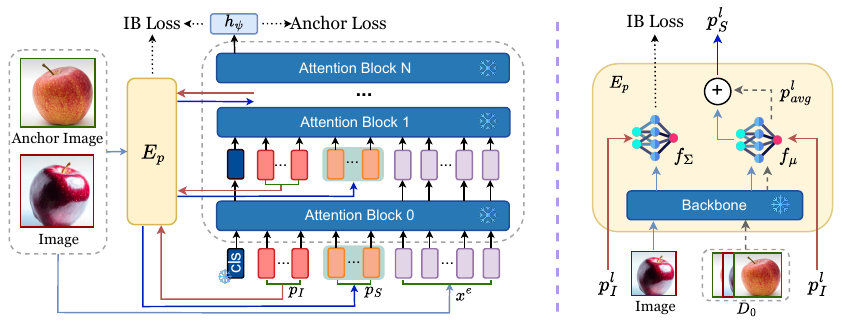}
    \caption{Overall training scheme of \ASP{} for the base task. For incremental tasks where $t>0$, \ASP{} updates only $\boldsymbol{p}_{avg}$ using~\cref{eq:EMA}. \textbf{Left:} The pre-trained backbone remains frozen during training and the prompts are inserted into layers between attention blocks. The \TIP{} $\boldsymbol{p}_I$ are initialized from an attention-aware aspect, while the \TSP{} $\boldsymbol{p}_S$ are derived from the prompt encoder $E_{\boldsymbol{p}}$. At the beginning of each training epoch, anchor images are selected using~\cref{eq:anchor_sample}. Throughout the training, the prompts and $E_{\boldsymbol{p}}$ are optimized using IB loss and Anchor loss as specified in~\cref{eq:overall_loss}. \textbf{Right:} Details of the prompt encoder $E_{\boldsymbol{p}}$. Image features extracted by the frozen pre-trained backbone are fed to two tiny networks $f_\mu$ and $f_\Sigma$. At the start of each training epoch, $p_{avg}$ is calculated via~\cref{eq:p_avg} using $\boldsymbol{p}_I$ and all data in the base task. Within a training epoch, the output of $f_\Sigma$ contributes to IB loss, while $\boldsymbol{p}_S$ results from blending $p_{avg}$ and the output of $f_\mu$, as outlined in~\cref{eq:ps}.}
    \label{fig:alg}
\end{figure*}

A base model with good generalization ability is beneficial for adapting to few-shot new classes~\cite{zhou2022forward, song2023learning}. To prevent overfitting on base classes after sufficient training, and to leverage the generalization ability of pre-trained ViT for learning new classes with limited data, \ASP{} fixes the pre-trained backbone and learns prompts that can transfer the knowledge learned from base classes to new classes. Inspired by DualP~\cite{wang2022dualprompt}, we decompose the prompts into attention-aware task-invariant prompts and self-adaptive task-specific prompts. In~\cref{sec:TIP}, \ASP{} maintains consistent attention across all task-invariant prompts for any given task, thereby containing minimal task-specific information. In~\cref{sec:TSP}, \ASP{} employs a prompt encoder to map an input image to \TSP{} leveraging the \textit{Information Bottleneck} theory, which has been proven to enhance generalization ability~\cite{lin2022bayesian}. Thus, the prompt encoder can also be used for new class data without further training. Finally,~\cref{sec:anchor} introduces an Anchor Loss, which enlarges class margins by pulling class features toward their class centers, thereby further improving model discrimination ability. \ASP{} only trains the model on the base task using sufficient data from base classes. Subsequently, the prompts are updated using~\cref{eq:EMA} for few-shot incremental tasks. Overall, our training scheme is shown in~\cref{fig:alg}.

\subsection{Attention-Aware Task-Invariant Prompts}
\label{sec:TIP}

 Similar to DualP~\cite{wang2022dualprompt}, task-invariant prompts are fixed after training on base classes and used for subsequent few-shot incremental tasks. Despite DualP employing the same task-invariant prompts for all tasks, it is not true that they convey identical information across different tasks due to the difference of attention on each prompt token. Consequently, these task-invariant prompts continue to offer task-specific information. To diminish the task-specific information stored in prompts, \ASP{} promotes consistent attention weight on each prompt token. The attention on different tokens can be measured by the attention matrix $A = {\rm{softmax}} (\frac{Q K^T}{\sqrt{d_k}})$. Denote the $i^{th}$ token in layer $\boldsymbol{h}^l$ as $\boldsymbol{t}_i \in \mathbb{R}^{1 \times D}$, the attention from the $i^{th}$ token to $j^{th}$ token is:
\begin{small}
\begin{equation}
\begin{split}
    A_{ij} = \frac{exp(\boldsymbol{t}_i W^Q \cdot \boldsymbol{t}_j W^K)}{\sum\nolimits_{m=1}^{1+L_{\boldsymbol{p}}+L_x} exp(\boldsymbol{t}_i W^Q \cdot \boldsymbol{t}_m W^K)} 
\end{split}
\end{equation}
\end{small}

\begin{figure*}[t]
    \centering
    \includegraphics[width=1\textwidth]{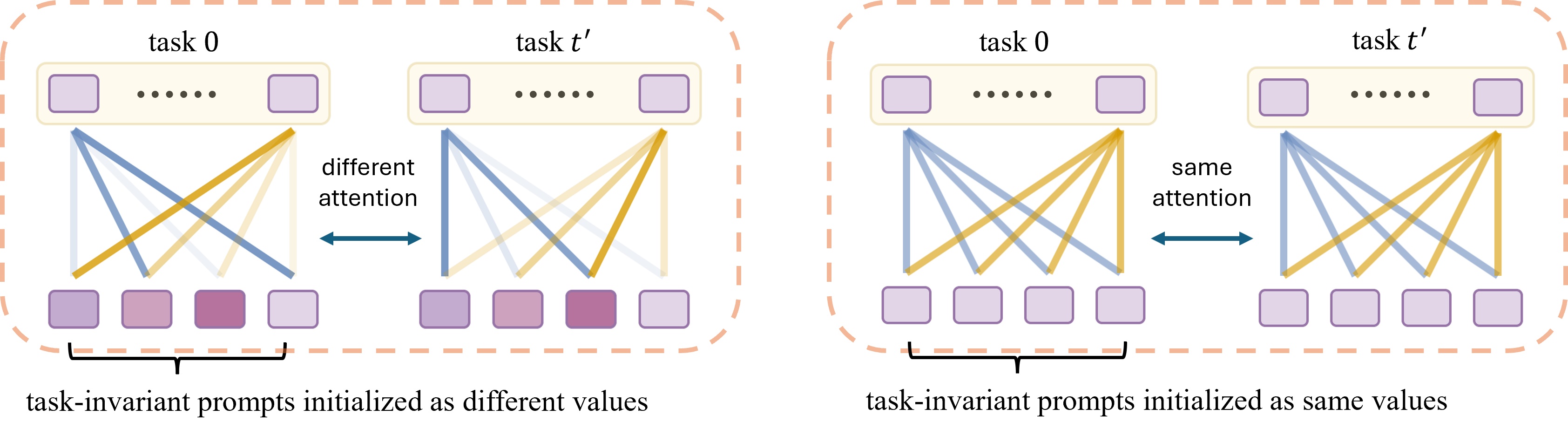}
    \caption{Attention on task-invariant prompts between different tasks. A deeper line color indicates greater attention. \textbf{Left:} When initialized with different values, attention on each prompt differs across tasks, thereby providing inconsistent information. \textbf{Right:} \ASP{} initializes \TIP{} with the same values, ensuring consistent attention across tasks and providing uniform information. }
    \label{fig:prompt}
\end{figure*}

\noindent
The attention is conditioned on the value of the prompt token, and the same attention can be guaranteed if two prompt tokens have the same values. The simplest way is initializing each prompt token with the same value before training, and the values will keep the sample during the model update using gradient decent algorithms~\cite{ruder2016overview}. In the base class training task, we use prompts $\boldsymbol{p}_I^l \in \mathbb{R}^{L_{\boldsymbol{p}_I} \times D}$ where each token is initialized using the same values as the task-invariant prompt. The comparison between our attention-aware \TIP{} and widely used randomly initialized prompt is shown in~\cref{fig:prompt}. As $\boldsymbol{p}_I^l$ always contribute the same to the model output regardless of task and image class, it is encouraged to contain knowledge shared among all base classes. This class-invariant knowledge can also be used for new classes in incremental tasks, thus it is also task-invariant knowledge. After training on base classes, the \TIP{} are fixed during few-shot incremental tasks.

\subsection{Self-Adaptive Task-Specific Prompts}
\label{sec:TSP}

However, relying solely on \TIP{} may lead to underfitting, as it only considers the shared attributes across different tasks and ignores the unique attributes. To incorporate task-specific information into prompts, prior studies~\cite{wang2022learning, wang2022dualprompt} introduce a key-query mechanism to generate task-specific prompts based on input images. However, these methods require a large amount of training data to capture the task-specific information and store it in prompts. In contrast, \ASP{} utilizes a compact neural network as a prompt encoder $E_{\boldsymbol{p}}$ to convert input images to task-specific prompts. This prompt encoder is initially trained with sufficient base class data to acquire encoding capabilities. To ensure these capabilities are extendable to new classes, \ASP{} enhances the generalization ability of the prompt encoder by adopting principles from the \textit{Information Bottleneck} theory~\cite{alemi2017deep}.

\noindent\textbf{Prompt Learning Objective:} Inspired by IB theory \cite{alemi2017deep}, we propose the following learning objective for prompt learning: 
\begin{align}
    \gL_{IB} 
    &= I(\mathcal{P}; \mathcal{X})  - \gamma I(\mathcal{P}; \mathcal{Y})  \label{eq:prompt}
\end{align}
We utilize the random variable $\mathcal{P}$ to represent the latent prompt associated with input $\mathcal{X}$. The mutual information between the latent prompt $\mathcal{P}$ and data labels $\mathcal{Y}$ is denoted as $I(\mathcal{P}; \mathcal{Y})$, while $I(\mathcal{P}; \mathcal{X})$ denotes the mutual information between latent prompt $\mathcal{P}$ and input data $\mathcal{X}$. Here, $\gamma$ is a constant. In~\cref{eq:prompt}, the term $I(\mathcal{P}; \mathcal{X}) - \gamma I(\mathcal{P}; \mathcal{Y})$ constitutes the \textit{Information Bottleneck} (IB) loss. Maximizing the mutual information between $\mathcal{P}$ and $\mathcal{Y}$ aims to strengthen the correlation between them, thereby capturing semantic knowledge. Conversely, minimizing the mutual information between $\mathcal{X}$ and $\mathcal{P}$ aims to mitigate the influence of extraneous information from input $\mathcal{X}$ on the prompt $\mathcal{P}$, thereby enhancing generalization. However, computing the mutual information is generally intractable since it often necessitates integration over the joint distribution of the involved variables. This integration poses computational or analytical challenges. Variational inference offers a framework for approximating these intractable integrals.
We formulate the variational lower bound as follows:
{
\small
\begin{align} \label{eq:variational_lower_bound}
   & I(\mathcal{P}; \mathcal{Y}) - \eta I(\mathcal{P}; \mathcal{X})  \geq \int P(\vx)P(\vy|\vx)P(\vp|\vx)\log P(\vy|\vp) d\vx d\vy d\vp \\ \nonumber
   &- \eta \int  P(\vx)P(\vp|\vx) \log \frac{P(\vp| \vx)}{r(\vp)}d\vx d\vp
\end{align}
}

\noindent
Here, $r(\vp)$ serves as a variational marginal approximation of the intractable marginal $P(\vp)$, and it is chosen as $r(\vp) = \mathcal{N}(0, I)$.
The detailed derivation steps are provided in the Appendix.

We now delve into the calculation of~\cref{eq:variational_lower_bound}. The term $P(\vy, \vx) = P(\vx)P(\vy|\vx)$ is approximated by the empirical data distribution
$P(\vx|\vy)=$ \\$\frac{1}{N} \sum_{n=1}^{N}\delta_{\vx_n}(\vx) \delta_{y_n}(\vy)$, where $\delta_{\vx_n}$ and $\delta_{y_n}$ represent the Dirac delta function. Assuming the prompt encoder $E_{\boldsymbol{p}}$ follows the form: 
$ P(\vp|\vx) = \mathcal{N}(\boldsymbol{p}|{f_\mu}(\vx), {f_\Sigma}(\vx))$
where two fully connected networks $f_\mu$ and $f_\Sigma$ outputs the $K$-dimensional mean $\mu$ of $\vp$ and the $K \times K$ covariance matrix $\Sigma$. Then, employing the reparameterization trick \cite{kingma2013auto,alemi2016deep}, we rewrite $P(\vy|\vx)d\vp$ as $P(\vepsilon)d\vepsilon$. By calculating the Kullback-Leibler (KL) divergence between $P(\vp|\vx)$ and $r(\vp)$, and combining all components, we obtain the empirical \textit{Information Bottleneck} loss function as described in~\cref{eq:idbound}, which we aim to minimize.
\begin{align} \label{eq:idbound}
    \gL_{IB} = \sE_{\vepsilon \sim \gN(0, I)}[- \log P(y| (\vx, \vepsilon))] + \KL(P(\vp|\vx)|r(\vp))
\end{align}
We use the prompt encoder $E_{\boldsymbol{p}}$ to convert an image to the corresponding \TSP{}. Inspired by previous works~\cite{wang2022learning, wang2022dualprompt}, $E_{\boldsymbol{p}}$ first uses the pre-trained backbone $f_\theta$ to extract the embedding features of input images and then fed to $f_\mu$ to obtain prompt features. Aiming to transfer the knowledge learned from base classes to new classes, the prompt encoder also takes \TIP{} as input to generate \TSP{}. To further improve the generalization ability of our \TSP{} for new classes, we consider all prompt features obtained from $\mathcal{\bm{X}}_0$ to provide general information together with the prompt features of an input image to construct the corresponding task-specific prompts $\boldsymbol{p}_S$. In this way, $\boldsymbol{p}_S$ can avoid overfitting to a single input image and become easier to generalize to unseen classes. During the base class training stage, the task-specific prompts for layer $l$ are given as:
\begin{small}
\begin{equation}
\begin{split} \label{eq:ps}
    \boldsymbol{p}_S^l = \alpha \boldsymbol{p}_{avg}^l + (1-\alpha) f_\mu([\boldsymbol{p}_I^l; f_\theta(\vx,\vepsilon)])
\end{split}
\end{equation}
\end{small}
\begin{small}
\begin{equation}
\begin{split} \label{eq:p_avg}
    \boldsymbol{p}_{avg}^l = \frac{1}{N_0} \sum_{N_0} f_\mu([\boldsymbol{p}_I^l; f_\theta(\vx_0)])
\end{split}
\end{equation}
\end{small}

\noindent
where $\alpha$ is a hyperparameter. $\boldsymbol{p}_{avg}^l$ is the average prompt features of data in the base task, which is recalculated at the beginning of each training epoch. For incremental task $t$, information of new classes is incorporated into $\boldsymbol{p}_{avg}^l$ using Exponential
Moving Average (EMA)~\cite{zhang2022grow}:
\begin{small}
\begin{equation}
\begin{split} \label{eq:EMA}
    \boldsymbol{p}_{avg}^{l} = \beta \boldsymbol{p}_{avg}^{l} + (1-\beta) \frac{1}{N_t} \sum_{N_t} f_\mu([\boldsymbol{p}_I^l; f_\theta(\vx_t)])
\end{split}
\end{equation}
\end{small}

\noindent
where $\beta$ is a hyperparameter to control the adapting speed. 
To balance the influence of task-invariant and task-specific prompts, their prompt length is set as the same, i.e. $L_{\boldsymbol{p}_S}=L_{\boldsymbol{p}_I}$. Finally, the prompts inserted into layer $l$ is:
\begin{small}
\begin{equation}
\begin{split}
    \boldsymbol{p}^l = [\boldsymbol{p}_I^l; \boldsymbol{p}_S^l]
\end{split}
\end{equation}
\end{small}

\subsection{Anchor Loss}
\label{sec:anchor}

During base classes training, we aim to obtain a feature extractor that can (1) maximize the distance between inter-class feature embeddings, and (2) minimize the distance between intra-class feature embeddings. Furthermore, we also want to obtain a classifier head $h_\psi$ that can accurately classify these features into class predictions. Following ~\cite{peng2022few, wang2023few}, we use a fully connected layer without bias term as the classifier head $h_\psi$, and the prediction is obtained by measuring the cosine similarity between feature embeddings and weights $W$ in the classifier head:
\begin{small}
\begin{equation}
\begin{split}
    y_k = \frac{W_k^T f_{\theta, \boldsymbol{p}}(\vx,\vepsilon)}{||W_k|| \cdot ||f_{\theta, \boldsymbol{p}}(\vx,\vepsilon)||}
\end{split}
\end{equation}
\end{small}

\noindent
where $||\cdot||$ is $l_2$ normalization. 
The weight $W_k$ can be seen as the prototype of class $k$ \cite{peng2022few, liudeja}. To separate class features and prototypes, the Cross-Entropy loss is used in~\cref{eq:idbound}:
\begin{align} \label{eq:loss_IB}
    \gL_{IB} = -\frac{1}{N} \sum_N log \frac{exp(y_k)}{\sum_{k' \in K} exp(y_{k'})} + \KL(P(\vp|\vx)|r(\vp))
\end{align}
 The Cross-Entropy loss simultaneously pulls class features $f_{\theta,\boldsymbol{p}}(\vx_k)$ to its class prototype $W_k$, and pushes far away from other class prototypes $W_{i \neq k}$. However, only the pull action is accurate while the push action may lead the class prototype to diverge from the class mean and cause misclassification. Similar to previous works~\cite{wang2023few, zhou2023revisiting}, we replace $W$ with class mean $\vc$ after training on base classes. Therefore, it is necessary to align class features with their class mean, which can also reserve places for new classes to improve new class accuracy~\cite{zhou2022forward, song2023learning}. For any input sample, we maximize the cosine similarity between its features and the corresponding class mean:
\begin{small}
\begin{equation}
\begin{split}
    \gL_c = 1 - \frac{\vc_k^T f_{\theta, \boldsymbol{p}}(\vx_k)}{||\vc_k|| \cdot ||f_{\theta, \boldsymbol{p}}(\vx_k)||}
\end{split}
\end{equation}
\end{small}

\noindent
As the class mean keeps changing during the training process, it is resource-consuming to compute the accurate class mean after each mini-batch. Thus, we use an anchor sample to estimate the class mean. At the beginning of each epoch, the accurate class mean is computed, and the sample that has the largest similarity with the class mean is selected as the anchor sample for that class:
\begin{small}
\begin{equation}
\begin{split} \label{eq:anchor_sample}
    \hat{\vx}_k = \arg \max_{\vx \in \mathcal{X}_k} \frac{\vc_k^T f_{\theta, \boldsymbol{p}}(\vx)}{||\vc_k|| \cdot ||f_{\theta, \boldsymbol{p}}(\vx)||}
\end{split}
\end{equation}
\end{small}

\noindent
Then the estimated class mean for $L_c$ is $\hat{\vc}_k = f_{\theta, \boldsymbol{p}}(\hat{\vx}_k)$. The training loss is:
\begin{small}
\begin{equation}
\begin{split} \label{eq:overall_loss}
    \gL = \gL_{IB} + \lambda \gL_c
\end{split}
\end{equation}
\end{small}

\noindent
where $\lambda$ is a hyperparameter.


\section{Experiments}
\label{experiments}

In this section, we first introduce the experiment details of FSCIL, including datasets, evaluation protocol, training details, and baseline methods. Subsequently, we compare \ASP{} with baselines on three benchmark datasets, which show the effectiveness of \ASP{}. In addition, the ablation study verifies the effectiveness of different components in \ASP{}. Lastly, we provide more experimental results for further analysis. We will release our code after the paper decision.

\subsection{Implementation Details}
\label{sec:implementation}

\textbf{Datasets:} Following FSCIL works~\cite{zhou2022forward, wang2023few} and prompt-based CIL works~\cite{wang2022dualprompt, wang2023hierarchical,sun2023pilot}, we evaluate the performance on CIFAR100~\cite{krizhevsky2009learning}, CUB200-2011~\cite{wah2011caltech}, and ImageNet-R~\cite{hendrycks2021many} dataset. Similar to prior studies~\cite{zhou2022forward, wang2023few}, the datasets are split to form the FSCIL tasks. In detail, CIFAR100 is divided into 60 base classes and 40 new classes. The new classes are further divided into eight $5$-way $5$-shot incremental tasks. CUB200 and ImageNet-R are divided into $100$ classes for the base task, and the left $100$ classes are divided into ten $10$-way $5$-shot incremental tasks.

\noindent\textbf{Evaluation protocol:} Following previous works~\cite{Tao_2020_CVPR, zhou2022forward, chi2022metafscil}, we denote the Top-1 accuracy on all seen tasks $0,...,t$ after the $t$-th task as $A_t$. The average accuracy $A_{avg}=\frac{\sum\nolimits_{t=0}^{T}A_t}{T+1}$ measures the overall performance during all incremental tasks. And the forgetting phenomenon is measured by the performance dropping rate (PD), i.e., $PD = A_0 - A_T$, where $0$ stands for the base task and $T$ stands for the last task. Furthermore, Harmonic Accuracy (HAcc)~\cite{peng2022few} is used to reflect the balanced performance across both base and new classes after task $T$: $A_h=\frac{2 \times A_o \times A_n}{A_o + A_n}$, where $A_o$ is the accuracy of base class in task $0$ and $A_n$ is the average accuracy of all classes in tasks $t>0$.

\noindent\textbf{Training details:} All experiments are conducted with PyTorch~\cite{paszke2019pytorch} on NVIDIA RTX A6000. We follow works~\cite{smith2023coda, zhou2023revisiting} to choose ViT-B/16-1K~\cite{dosovitskiy2020image}, which pre-trained on ImageNet1K as the backbone $f_\theta$. For all datasets, the input images are resized to $224 \times 224$ and trained 20 epochs using SGD optimizer. The learning rate is set as 0.01 for CIFAR100 and CUB200 while 0.03 is used for ImageNet-R. We use batch size of 48 for CIFAR100 and batch size of 24 for both CUB200 and ImageNet-R. The prompt token length $L_{p_g}=L_{p_d}$ of $10$ is used for ImageNet-R, and $3$ for CIFAR100 and CUB200. All experiments are run using three random seeds and the average results are reported.

\noindent\textbf{Baselines:} We first compare two widely used CIL methods iCaRL~\cite{rebuffi2017icarl} and Foster~\cite{wang2022foster}. Besides, we also compare the SOTA FSCIL algorithms: CEC~\cite{Zhang_2021_CVPR}, FACT~\cite{zhou2022forward} and TEEN~\cite{wang2023few}. Lastly, we compare the recent SOTA prompt-based CIL approaches: L2P~\cite{wang2022learning}, DualP~\cite{wang2022dualprompt}, and CodaP~\cite{smith2023coda}.


\subsection{Benchmark Comparisons}
\label{main_exp}

\begin{figure*}
    \centering
    \includegraphics[width=1\textwidth]{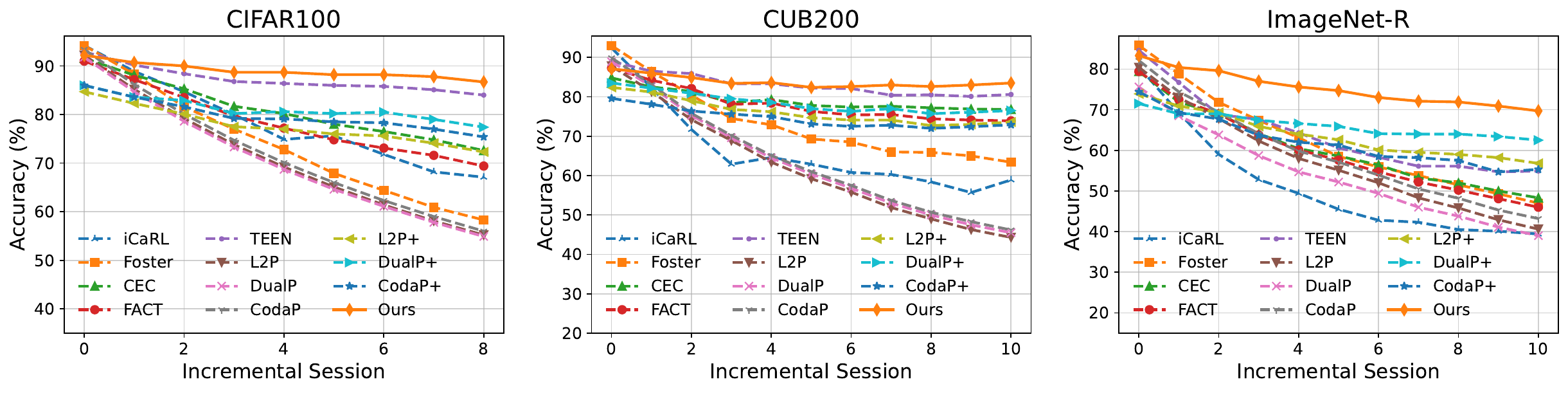}
    \caption{Detailed Top-1 accuracy $A_t$ in each incremental task on three benchmark datasets. \ASP{} outperforms baselines in most tasks.}
    \label{fig:main}
\end{figure*}

In this section, we report the performance of baselines and \ASP{} under FSCIL setting. For CIFAR100 and ImageNet-R, the detailed accuracy in each task and the three evaluation metrics are shown in~\cref{tab:cifar} and~\cref{tab:inr}. The detailed performance of CUB200 can be found in the appendix. Furthermore, the performance curves of the Top-1 accuracy $A_{t}$ in each incremental task are shown in~\cref{fig:main}.

Based on the experimental results on CIFAR100, CUB200 and ImageNet-R, \ASP{} achieves the best Top-1 accuracy $A_T$ of $86.7\%$, $83.5\%$ and $69.7\%$, surpassing the second best by $2.7\%$, $2.9\%$ and $7.2\%$ respectively. Furthermore, \ASP{} usually performs the best on $A_t$ before the last task, except the first task. In the first task, classical CIL and FSCIL methods fine-tune all the parameters using sufficient data, thus it is reasonable that they perform better than the prompt-based CIL approaches. Besides, \ASP{} performs the best on the average accuracy $A_{avg}$, which achieves $89.0\%$, $83.8\%$ and $75.3\%$ and surpasses the second best by $1.7\%$, $0.7\%$ and $9.2\%$ on three benchmark datasets respectively. In addition, \ASP{} achieves the lowest PD on CIFAR100 and CUB200 while achieving the second lowest on ImageNet-R. For the HAcc metric, \ASP{} also achieves the best of $85.3\%$, $83.4\%$ and $67.0\%$ on three benchmark datasets, which can demonstrate the superiority of \ASP{} in incrementally learning new classes while preserving performance on base classes at the same time.

Interestingly, we find that prompt-based CIL approaches learn nearly nothing about the new classes based on the HAcc metric. We think the main reason may be the severe overfitting of few-shot data in the classifier head, which is a fully connected layer. Thus, we replace the original classifier head with the class mean to form a prototypical network, which is denoted as L2P+, DualP+, and CodaP+. The HAcc results in~\cref{tab:cifar,tab:inr} show that this modification largely improves their ability to learn new tasks under the FSCIL setting.

\begin{table*}[tb]
\small
\centering
\caption{Detailed Top-1 accuracy $A_t$ in each incremental task, average accuracy $A_{avg}$, performance dropping rate (PD) and Harmonic Accuracy (HAcc) on CIFAR100 dataset. $\uparrow$ means higher is better, and $\downarrow$ means lower is better. }

\label{tab:cifar}
\begin{tabular}{lcccccccccccc}
\toprule
\multicolumn{1}{l}{\multirow{2}{*}{Method}} & \multicolumn{9}{c}{Accuracy $A_t$ in each task (\%) $\uparrow$} & \multirow{2}{*}{$A_{avg} \uparrow$} & \multirow{2}{*}{PD$ \downarrow$} & \multirow{2}{*}{HAcc$\uparrow$} \\ \cline{2-10} 
 & 0 & 1 & 2 & 3 & 4 & 5 & 6 & 7 & 8 && \\ \hline 
iCaRL  &\textbf{94.2} & 88.9 & 84.7 & 80.0 & 74.9 & 75.6 & 71.8 & 68.2 & 67.1 &78.4 & 27.2 & 57.5 \\
Foster  &\textbf{94.2} & 88.3 & 81.6 & 77.0 & 72.8 & 67.9 & 64.4 & 60.9 & 58.3 &73.9 & 35.9 & 11.0   \\ \midrule
CEC    &91.6 & 88.1 & 85.3 & 81.7 & 80.2 & 78.0 & 76.5 & 74.8 & 72.6 &81.0 & 19.0 & 64.1   \\
FACT   &91.0 & 87.2 & 83.5 & 79.7 & 77.2 & 74.8 & 73.1 & 71.6 & 69.4 &78.6 & 21.7 & 55.5   \\
TEEN   &92.9 & 90.2 & 88.4 & 86.8 & 86.4 & 86.0 & 85.8 & 85.1 & 84.0 &87.3 & 8.8 & 81.2   \\ \midrule
L2P &92.2 & 85.2 & 79.2 & 73.8 & 69.2 & 65.1 & 61.4 & 58.1 & 55.2 &71.1 & 37.0 & 0.0 \\
DualP &91.8 & 84.7 & 78.6 & 73.3 & 68.7 & 64.6 & 61.1 & 57.8 & 54.9 &70.6 & 36.9 & 0.1 \\
CodaP &93.4 & 86.2 & 80.1 & 74.7 & 70.1 & 66.0 & 62.3 & 59.0 & 56.0 &72.0 & 37.4 & 0.0 \\
L2P+    &84.7 & 82.3 & 80.1 & 77.5 & 77.0 & 76.0 & 75.6 & 74.1 & 72.3 &77.7 & 12.4 & 68.0  \\
DualP+  &86.0 & 83.6 & 82.9 & 80.2 & 80.6 & 80.2 & 80.5 & 79.0 & 77.4 &81.1 & 8.5 & 75.3  \\
CodaP+  &86.0 &83.6 &81.6 &79.2 &79.1 &78.5 &78.3 &77.0 &75.4 &79.9 &10.6 &72.2 \\ \midrule
\textbf{Ours}  &92.2 & \textbf{90.7} & \textbf{90.0} & \textbf{88.7} & \textbf{88.7} & \textbf{88.2} & \textbf{88.2} & \textbf{87.8} & \textbf{86.7} &\textbf{89.0} & \textbf{5.5} & \textbf{85.3}  \\
\bottomrule
\end{tabular}

\end{table*}

\begin{table*}[t]
\centering
\caption{Detailed Top-1 accuracy $A_t$ in each incremental task, average accuracy $A_{avg}$, performance dropping rate (PD) and Harmonic Accuracy (HAcc)      on ImageNet-R dataset. $\uparrow$ means higher is better, and $\downarrow$ means lower is better.}
\label{tab:inr}

 \scalebox{0.9}{
\begin{tabular}{lcccccccccccccc}
\toprule
\multicolumn{1}{l}{\multirow{2}{*}{Method}} & \multicolumn{11}{c}{Accuracy $A_t$ in each task (\%) $\uparrow$} & \multirow{2}{*}{$A_{avg} \uparrow$} & \multirow{2}{*}{PD$ \downarrow$} & \multirow{2}{*}{HAcc$\uparrow$} \\ \cline{2-12} 
\multicolumn{1}{c}{} & 0   & 1      & 2      & 3    & 4     & 5  & 6     & 7      & 8     &9     &10    &     & &   \\ \midrule			
iCaRL  &80.6 & 69.2 & 59.0 & 52.8 & 49.4 & 45.5 & 42.8 & 42.3 & 40.5 & 40.1 & 39.4 &51.0 & 41.2 & 36.1 \\
Foster  &\textbf{85.8} & 78.8 & 71.8 & 67.4 & 63.1 & 58.5 & 55.9 & 53.8 & 51.5 & 49.3 & 47.0 &62.1 & 38.7 & 36.8   \\ \midrule
CEC    &79.4 & 71.9 & 69.0 & 64.1 & 60.4 & 58.6 & 56.4 & 53.2 & 52.0 & 50.0 & 48.3 &60.3 & 31.1 & 32.6   \\
FACT   &79.4 & 72.5 & 69.0 & 63.8 & 60.1 & 57.6 & 54.7 & 52.2 & 50.2 & 48.1 & 46.0 &59.4 & 33.4 & 22.3   \\
TEEN   &84.6 & 76.7 & 68.8 & 67.6 & 64.3 & 60.6 & 58.3 & 56.1 & 56.1 & 54.7 & 54.9 &63.9 & 29.7 & 45.4   \\ \midrule
L2P &80.4 & 73.0 & 67.8 & 62.3 & 58.0 & 55.1 & 52.0 & 48.3 & 45.8 & 42.9 & 40.6 &56.9 & 39.8 & 1.0 \\
DualP &75.6 & 68.5 & 63.8 & 58.7 & 54.7 & 52.2 & 49.4 & 46.0 & 43.8 & 41.1 & 39.0 &53.9 & 36.7 & 2.7 \\
CodaP &82.1 & 74.4 & 69.4 & 64.1 & 59.8 & 57.1 & 53.9 & 50.5 & 48.2 & 45.3 & 43.2 &58.9 & 38.9 & 6.5 \\
L2P+    &73.9 & 70.9 & 69.3 & 65.9 & 64.0 & 62.6 & 60.1 & 59.5 & 59.0 & 58.2 & 56.8 &63.7 & 17.2 & 52.0   \\
DualP+  &71.5 & 69.0 & 69.0 & 67.4 & 66.6 & 65.9 & 64.1 & 64.0 & 64.0 & 63.4 & 62.5 &66.1 & \textbf{9.0} & 61.6  \\
CodaP+  &74.4 &69.3 &67.7 &63.7 &62.0 &61.3 &58.5 &58.2 &57.5 &54.7 &55.3 &62.0 &19.1 &54.5 \\ \midrule
\textbf{Ours}  &83.3 & \textbf{80.4} & \textbf{79.6} & \textbf{77.0} & \textbf{75.6} & \textbf{74.7} & \textbf{73.0} & \textbf{72.1} & \textbf{71.9} & \textbf{70.9} & \textbf{69.7} &\textbf{75.3} & 13.5 & \textbf{67.0}  \\
\bottomrule
\end{tabular}
}
\end{table*}

\subsection{Ablation Study}
\label{sec:abaltion}

We conduct ablation study on three benchmark datasets, and the implementation details are the same as the setting in~\cref{sec:implementation}. To verify the effectiveness of each component in \ASP{}, we alternatively remove each component from the framework and measure the average accuracy $A_{avg}$. The results are shown in~\cref{tab:ablation}, where \emph{ours w/o TIP} refers to removing the attention-aware task-invariant prompt $p_I$ from our framework, \emph{ours w/o TSP} refers to removing the self-adaptive task-specific prompt $p_S$ from our framework, \emph{ours w/o $\gL_c$} refers to removing the anchor loss $\gL_c$ from our framework. Finally, \emph{ours w/ Diff TIP} means we use a Gaussian distribution to initialize different values for each task-invariant prompt. The results show that removing any component leads to a performance drop in the average accuracy $A_{avg}$, which demonstrates the effectiveness of our design.

\begin{table}[t]
    \begin{minipage}{.48\textwidth}
        \centering 
        \caption{Ablation study of removing each component from \ASP{} respectively. The average accuracy $A_{avg}$ is reported on three benchmark datasets.}
            \scalebox{0.8}{
            \begin{tabular}{lccc}
            \toprule
            \multirow{2}{*}{Methods} & \multicolumn{3}{c}{$A_{avg}$}    \\ \cmidrule{2-4}
            & CIFAR100 & CUB200 & ImageNet-R \\\midrule
            Ours w/o TIP  &88.5 &83.4  &72.2 \\
            Ours w/o TSP  &88.1  &83.0  &73.9  \\
            Ours w/o $\gL_c$  &88.6  &83.2  &74.5 \\ 
            Ours w/ Diff TIP  &87.6  &82.6 &73.3  \\ \midrule
            Ours &\textbf{89.0}  &\textbf{83.8} &\textbf{75.3} \\
            \bottomrule
            \end{tabular}
            }
            \label{tab:ablation}
    \end{minipage}
    \hfill
    \begin{minipage}{.48\textwidth}
    \captionof{table}{Influence of the Incremental Shot on average accuracy $A_{avg}$ on three benchmark datasets.}
        \centering 
        \scalebox{0.8}{
        \begin{tabular}{lccc}
        \toprule
        \multirow{2}{*}{Methods} & \multicolumn{3}{c}{$A_{avg}$}    \\ \cmidrule{2-4}
        & CIFAR100 & CUB200 & ImageNet-R \\\midrule
        $1$-shot  &82.9 &78.6 &69.3 \\
        $5$-shot  &89.0 &83.8 &75.3  \\
        $10$-shot  &90.0 &85.3 &76.6 \\ 
        $20$-shot  &90.3 &85.8 &77.4  \\ 
        \bottomrule
        \end{tabular}
        }
        \label{table:shot}
    \end{minipage}
\end{table}


\subsection{Further Analysis}

\begin{figure}[ht]
\begin{center}
\centerline{\includegraphics[width=\columnwidth]{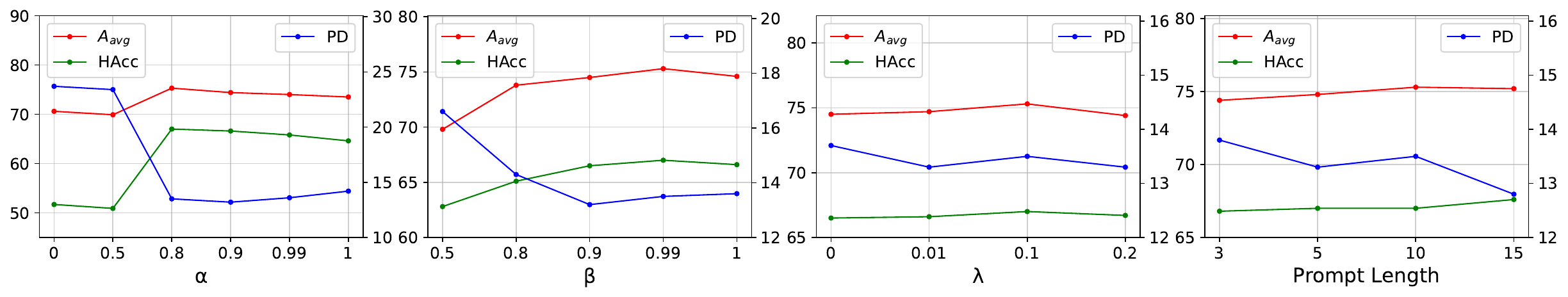}}
\caption{Sensitive analysis of $\alpha$, $\beta$, $\lambda$ and prompt length $L_g=L_d$. The average accuracy $A_{avg}$ is reported on ImageNet-R datasets. }
\label{fig:sensitive}
\end{center}
\end{figure}

\textbf{Hyper-Parameter:} We conduct sensitive analysis of $\alpha$, $\beta$, $\lambda$ and prompt length $L_g=L_d$. The same experimental setting of $10$-way $5$-shot is used on the ImageNet-R dataset and the results are shown in~\cref{fig:sensitive}. To achieve the highest $A_{avg}$, we choose $\alpha=0.8$, $\beta=0.99$ and $\lambda=0.1$ for all benchmark datasets. The prompt length is set as $3$ for CIFAR100 and CUB200 while is set as $10$ for ImageNet-R.

\noindent\textbf{Incremental Shot:} \ASP{} learns new classes using $K$-shot data, and we change the shot to find out the data influence on the average accuracy $A_{avg}$. The experimental setting is the same with~\cref{main_exp} and $1$, $10$, $20$-shot of new classes are provided in incremental tasks on three datasets. The results in~\cref{table:shot} show that more available samples per class in incremental tasks can help improve the model performance. However, we also find that the performance improvement becomes smaller as we increase shot $K$.

\noindent\textbf{Task-Specific Prompts:} The effectiveness of the self-adaptive task-specific prompts module is validated in~\cref{sec:abaltion}, and we provide further analysis to evaluate the importance of the average prompt features $p_{avg}$ in~\cref{eq:p_avg} and the EMA in~\cref{eq:EMA}. We remove $p_{avg}$ by setting $\alpha=0$ and don't update $p_{avg}$ for new tasks by setting the EMA parameter $\beta$ as $1$. Based on the results in~\cref{fig:sensitive}, both cases lead to performance drop on $A_{avg}$, PD and HACC, which validates the effectiveness of $p_{avg}$ and EMA.

\begin{figure}[ht]
\begin{center}
\centerline{\includegraphics[width=\columnwidth]{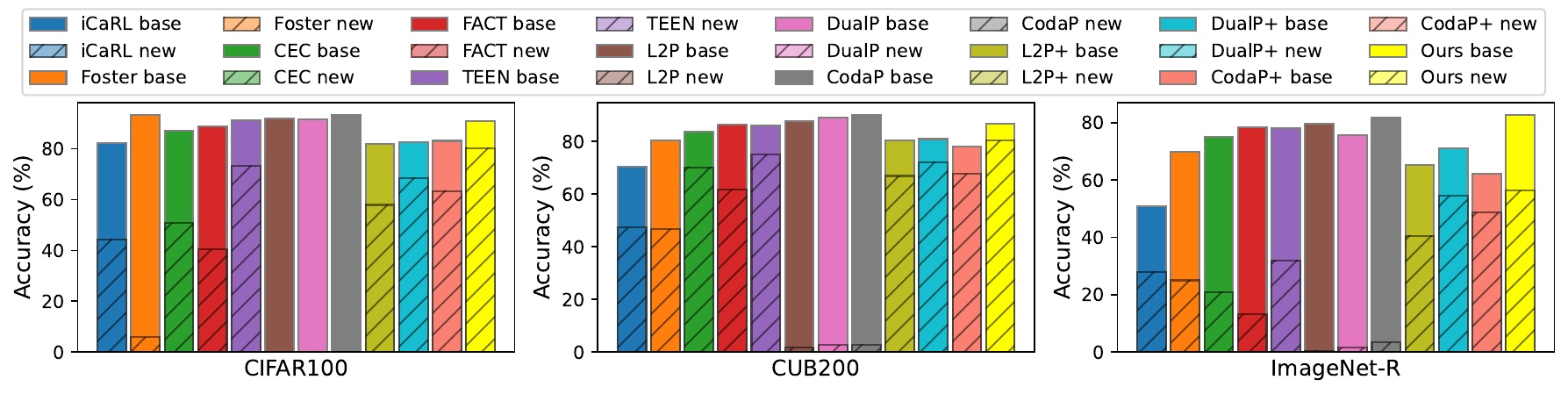}}
\caption{Comparison of baselines and \ASP{} on detailed accuracy of base and new classes after the last task. \ASP{} outperforms all baselines in terms of learning new classes while achieving competitive performance in maintaining performance on base classes.}
\label{fig:old_new}
\end{center}
\end{figure}

\noindent\textbf{Base\&New Class Accuracy:} Except HAcc metrics, we provide the detailed accuracy of the base classes from task $0$ and the new classes from task $t>0$ after the last task $T$. The results are shown in~\cref{fig:old_new}. \ASP{} outperforms all baselines in terms of learning new classes using few-shot data while achieving competitive performance in maintaining performance on base classes.


\section{Conclusion}
\label{conclusion}

In this work, we first point out the limitations of applying existing FSCIL methods and prompt-based CIL methods on FSCIL scenarios with large vision models. We further propose a new framework called \ASP{} to learn generalized prompts to leverage the generalization of pre-trained ViT for incrementally learning new classes with limited data. Under the FSCIL setting, \ASP{} outperforms baseline methods from classical CIL, FSCIL, and prompt-based CIL on three benchmark datasets.

\section*{Acknowledgments}

This work was partially supported by NSF IIS 2347592, 2347604, 2348159, 2348169, DBI 2405416, CCF 2348306, CNS 2347617.


%
%
\bibliographystyle{splncs04}
\bibliography{main}
\end{document}